\documentclass{article}

\usepackage{arxiv}

\usepackage[utf8]{inputenc} 
\usepackage[T1]{fontenc}    
\usepackage{hyperref}       
\usepackage{amsmath,amsfonts,amsthm} 

\DeclareMathOperator*{\argmin}{argmin}
\usepackage{url}            
\usepackage{booktabs}       
\usepackage{amsfonts}       
\usepackage{nicefrac}       
\usepackage{microtype}      
\usepackage{cleveref}       
\usepackage{lipsum}         
\usepackage{graphicx}
\usepackage{subcaption}
\graphicspath{{./figures/}}
\usepackage{svg} 
\usepackage{doi}
\usepackage[sorting=nyt,style=apa,backend=biber]{biblatex}
\DeclareLanguageMapping{american}{american-apa}
\title{Advanced For-Loop for QML algorithm search}

\date{}

\newif\ifuniqueAffiliation
\uniqueAffiliationtrue

\ifuniqueAffiliation 
\author{\href{https://orcid.org/0000-0002-1255-2561}{\includegraphics[scale=0.06]{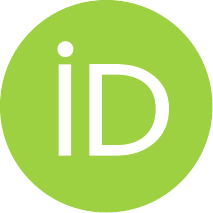}\hspace{1mm}FuTe~Wong}\thanks{} \\
  Institute of Medical Science, University of Toronto\\
	Vector Institute\\
	Department of Computer Science, University of Toronto\\
	\texttt{fute@cs.toronto.edu} \\
}
\else
\usepackage{authblk}

\newbox{\orcid}\sbox{\orcid}{\includegraphics[scale=0.06]{orcid.pdf}} 
\author[1,2,3]{%
	\href{https://orcid.org/0000-0002-1255-2561}{\usebox{\orcid}\hspace{1mm}FuTe~Wong\thanks{\texttt{fute@cs.toronto.edu}}}%
}
\affil[1]{Department of Electrical Engineering, University of Toronto, Canada}
\affil[2]{Vector Institute, Toronto, Canada}
\affil[3]{Department of Computer Science, University of Toronto, Toronto, Canada}
\fi

\renewcommand{\headeright}{}
\renewcommand{\undertitle}{A primer}

\hypersetup{
pdftitle={For-Loop},
pdfauthor={FuTe~Wong},
pdfkeywords={Agentic framework, Quantum Machine Learning, Machine Learning, LLM, Reasoning},
}

\begin{document}
\maketitle

\begin{abstract}
This paper introduces an advanced framework leveraging Large Language Model-based Multi-Agent Systems (LLMMA) for the automated search and optimization of Quantum Machine Learning (QML) algorithms. Inspired by Google DeepMind’s FunSearch, the proposed system works on abstract level to iteratively generates and refines quantum transformations of classical machine learning algorithms (concepts), such as the Multi-Layer Perceptron, forward-forward and backpropagation algorithms. As a proof of concept, this work highlights the potential of agentic frameworks to systematically explore classical machine learning concepts and adapt them for quantum computing, paving the way for efficient and automated development of QML algorithms. Future directions include incorporating planning mechanisms and optimizing strategy in the search space for broader applications in quantum-enhanced machine learning.
\end{abstract}


\section{Introduction}
\label{sec:headings1}
\begin{figure}[h]
	\centering
	\includegraphics[width=0.8\linewidth]{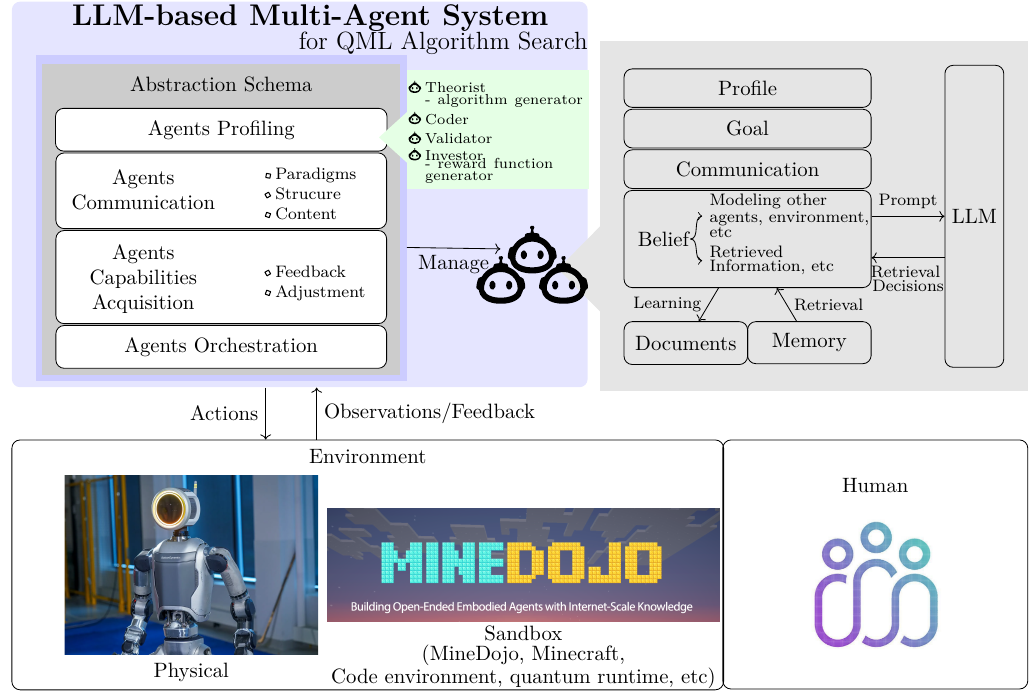}
  \caption{The architecture of the LLM-based Multi-Agent systems for QML search with dynamic role updating. (Figure adapted form \cite{guoLargeLanguageModel2024})}
	\label{fig:llmma}
\end{figure}

\indent Large Language Models (LLMs) have recently shown remarkable potential in reasoning and planning capabilities across a wide array of tasks \parencites{yaoTreeThoughtsDeliberate2023,ahnLargeLanguageModels2024}. Leveraging on a single LLM-based agent, LLM-based Multi-Agents (LLMMA) utilizing collective intelligence and specialized persona with differentiated skills offer advanced capabilities \parencites{wangRethinkingBoundsLLM2024,hongMetaGPTMetaProgramming2023,qianChatDevCommunicativeAgents2024,mandiRoCoDialecticMultiRobot2023,zhangBuildingCooperativeEmbodied2024}. Recently, LLMs have also risen to prominence in scientific discovery for their expansive knowledge bases, advanced reasoning capabilities, and human-friendly natural language interface \parencite{ai4scienceImpactLargeLanguage2023}. Inspired by Google DeepMind's FunSearch \parencite{romera-paredesMathematicalDiscoveriesProgram2024}, which looking for solutions for algorithmic problems via iteratively program code generation from LLM, we adapted a LLM-based Multi-Agents system for Quantum Machine Learning algorithms search (Figure \ref{fig:llmma}). Rather than studies like Quantum Architecture Search (QAS) in the gate sets of Parametrized Quantum Circuits \parencites[PQC;][]{wuQuantumDARTSDifferentiableQuantum2023,duQuantumCircuitArchitecture2022,nakajiGenerativeQuantumEigensolver2024}, search in program coding space allows agents to build more expressive resultant model architecture. Compared with a single LLM optimizer for neural architecture search \parencites[NAS;][]{zhangUsingLargeLanguage2023,yangLargeLanguageModels2024}, our approach extends agents with the capacities of tool use, memory via retrieval-augmented generation \parencite[RAG;][]{yanCorrectiveRetrievalAugmented2024}, and optimization techniques like multi-agent in-context learning \parencite{dongSurveyIncontextLearning2024}, direct preference optimization \parencite[DPO;][]{rafailovDirectPreferenceOptimization2023}, evolutionary algorithm \parencite{vansteinLLaMEALargeLanguage2024}, or feature steering \parencite{templeton2024scaling}. We construct the LLMMA QML algorithm search in a way that given a classical deep learning algorithm, such as forward-forward algorithm or backpropagation algorithm as described in the following parts, the system can facilitate the development workflow to find its optimized quantum counterpart.  

\section{Can LLM-based Agent find quantum machine learning algorithm from its classical counterpart?}
\label{sec:headings2}
\begin{figure}[h]
	\centering
	\includegraphics[width=0.7\linewidth]{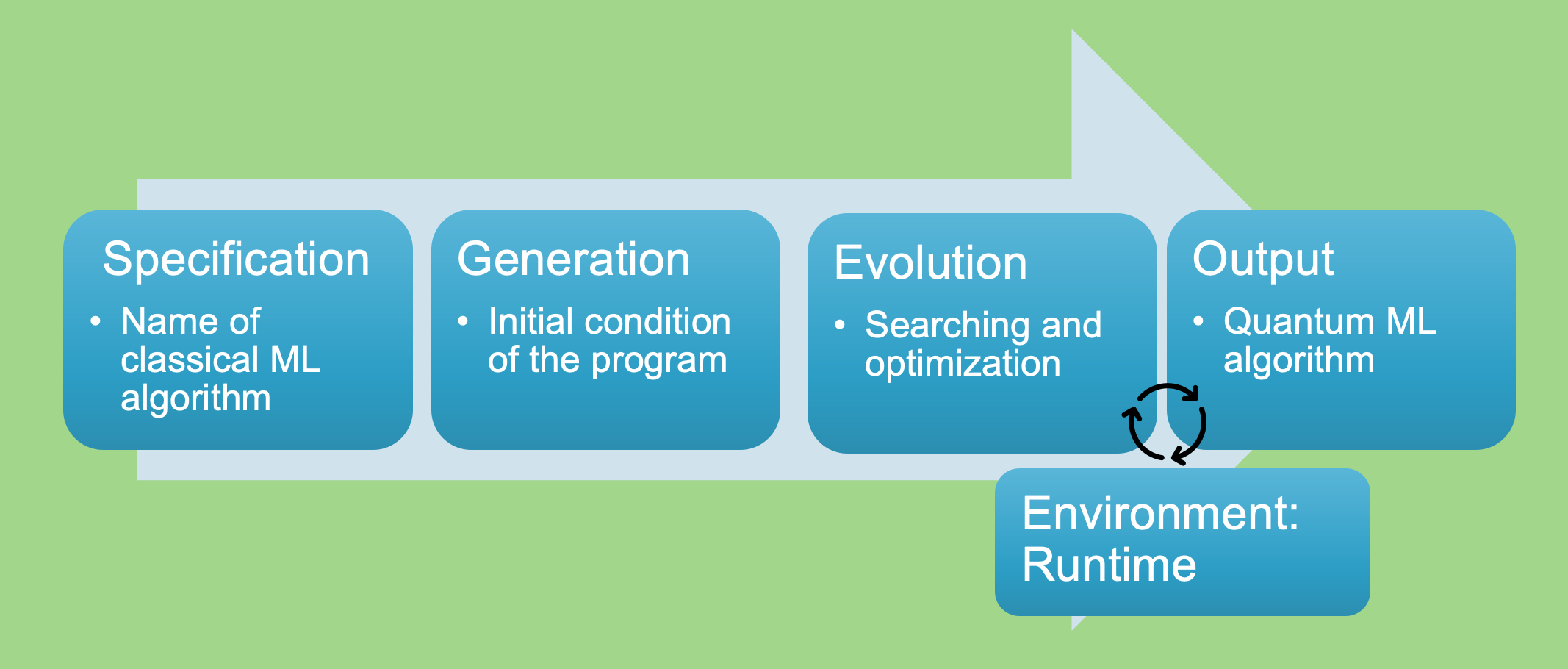}
	\caption{Flow chart of QML searching process.}
	\label{fig:fig2}
\end{figure}

Given the name of the classical algorithm, we have the agentic system is to generate the program code of the algorithm as initial condition, then conducting the evolution loop of searching and optimization to find its quantum counterpart (Figure \ref{fig:fig2})

\subsection{Multi-Layer Perceptron}
Firstly, we did an experiment on classical multilayer perceptron as given input to the agentic system. For example, given specification of classical machine learning algorithm, which is Multi-Layer Perceptron (MLP) with the following sample snippet:
\begin{figure}[h]
	\centering
	\includegraphics[width=0.7\linewidth]{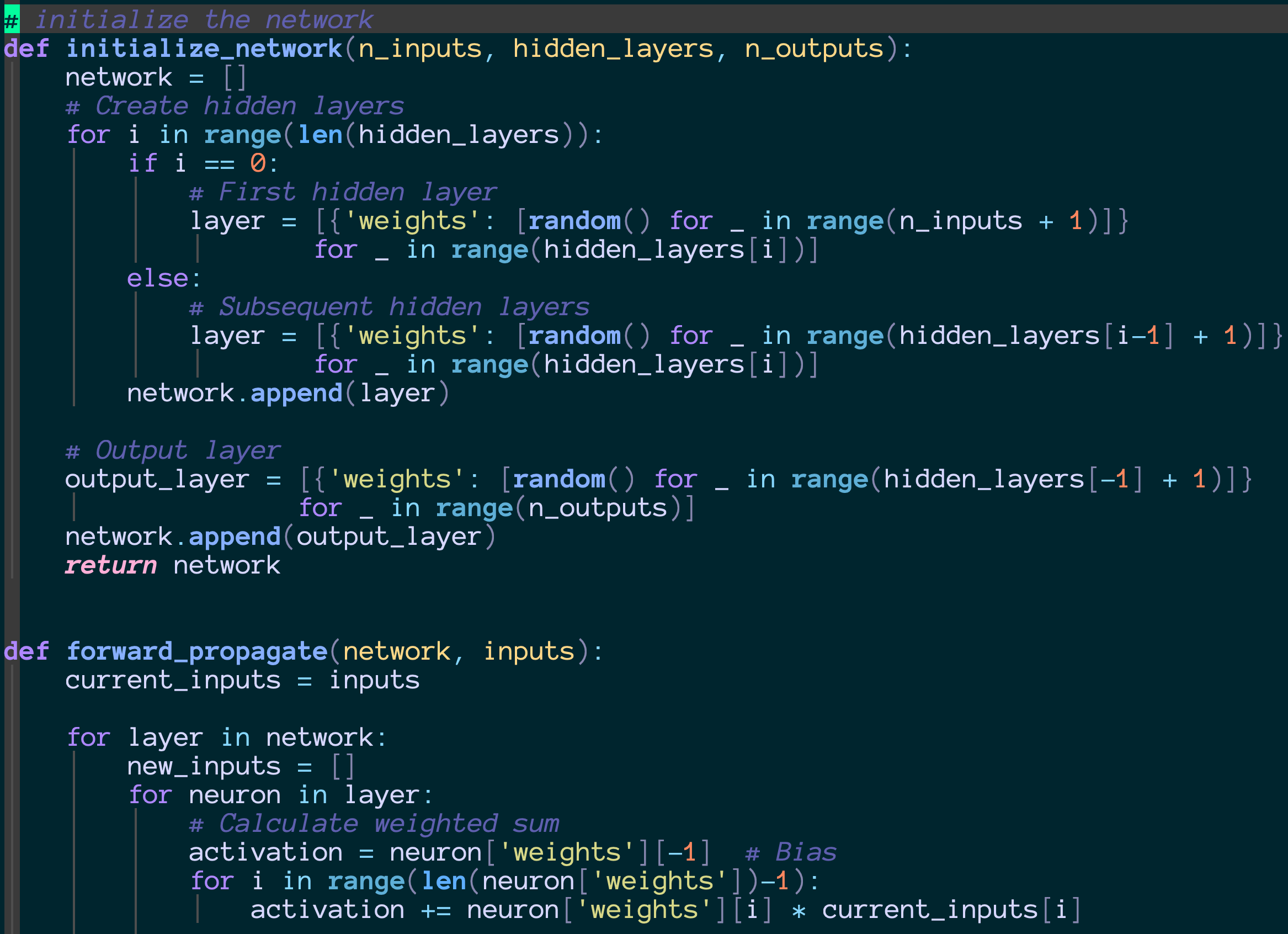}
	\caption{A snippet of Python implementation of MLP}
	\label{fig:fig3}
\end{figure}

The system would generate an implementation of quantum-enhanced version of the multi-layer perceptron and training the model. The following figure shows the quantum counter part and a piece of training dynamic that demonstrated that the model is learning during the training steps.

\begin{figure}[h]
  \begin{subfigure}[b]{0.5\linewidth}
		\includegraphics[width=1.2\linewidth]{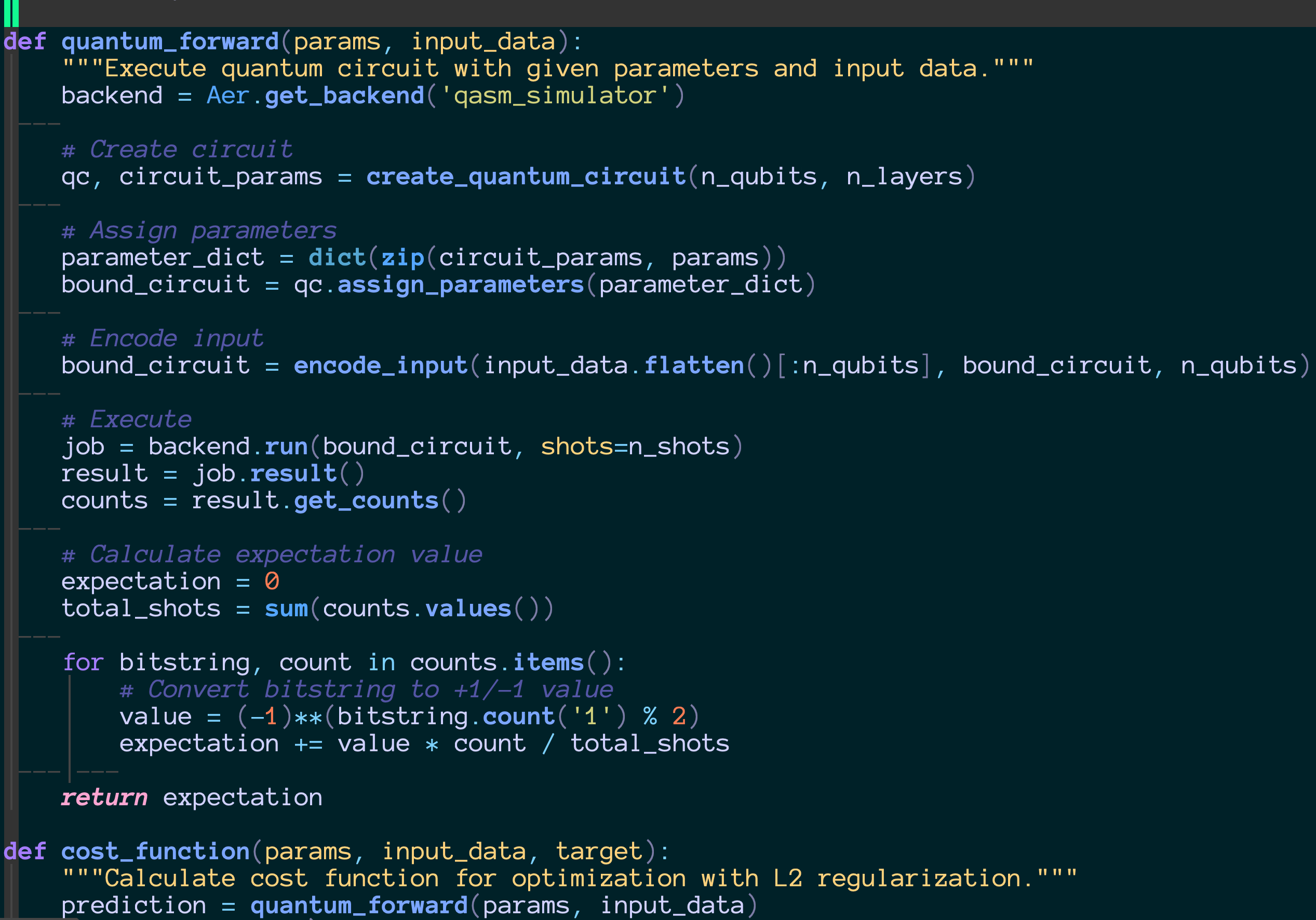}
	  \caption{A snippet of Python implementation of Quantum MLP}
	  \label{fig:fig4-1}
	\end{subfigure}
	 \hfill
	\begin{subfigure}[b]{0.4\linewidth}
		~~\includegraphics[width=1.1\linewidth]{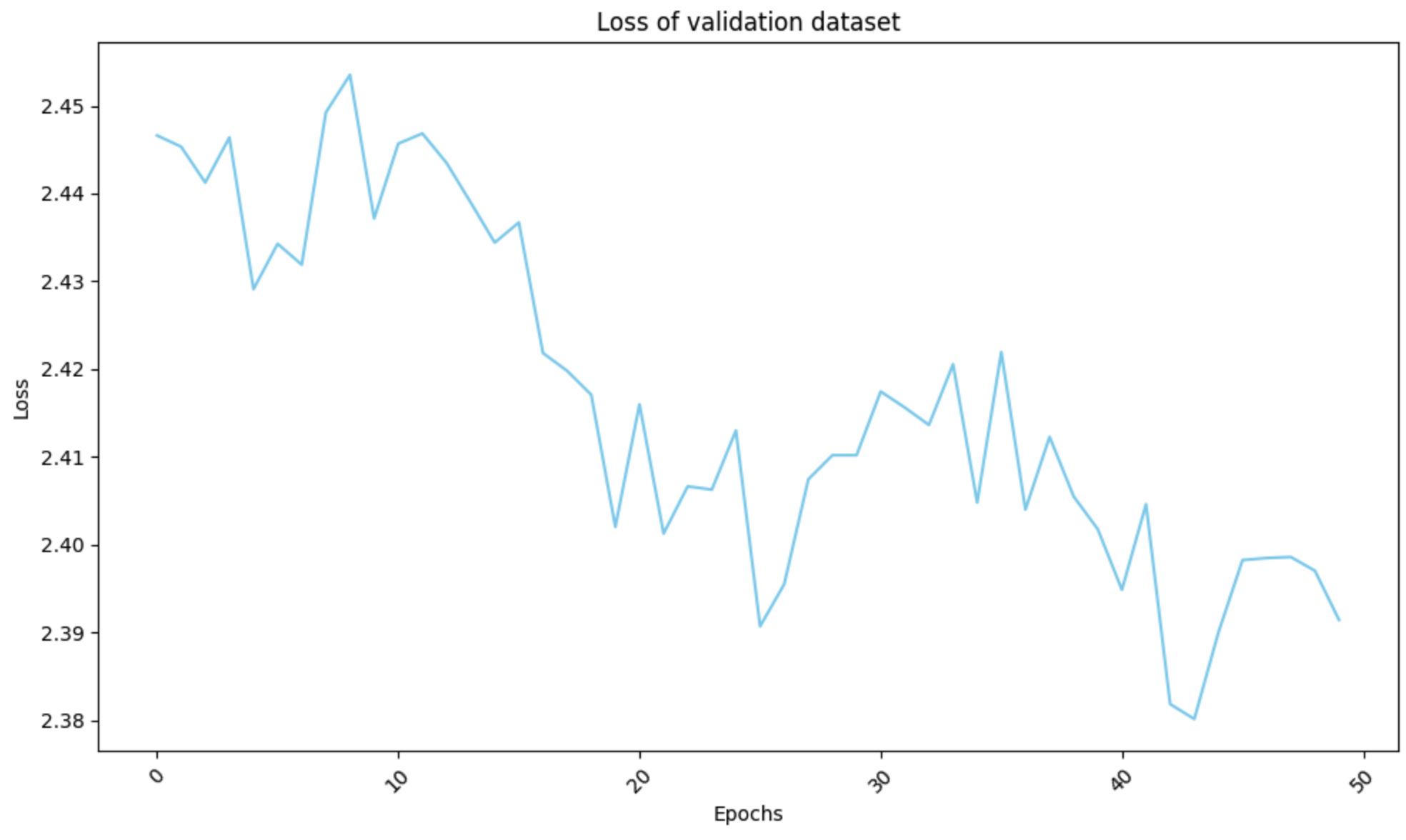}
	  \caption{Learning dynamics of the Quantum MLP}
	  \label{fig:fig4-2}
	\end{subfigure}
  \label{fig:fig4}
	\caption{}
  \end{figure}

\subsection{Forward-Forward Algorithm}
The initial motivation for the development of the forward-forward algorithm \parencite{hintonForwardForwardAlgorithmPreliminary2022} is the reflection of the biological implausibility of backpropagation. Although the backpropagation algorithm is a fundamental and powerful learning procedure in the machine learning field, there is no convincing evidence that the cortex of the human brain explicitly propagates error derivatives or stores neural activities for use in a backward propagation of error information.
To approximate how the human cortex learns, the forward-forward algorithm replaces the forward and backward passes of backpropagation with two forward passes that operate in the same way as each other, and applies a contrast learning scheme with different data with opposite objectives.
For example, the positive pass operates on real data and adjusts the weights to increase the goodness in every hidden layer. The negative pass operates on ``negative data'' and adjusts the weights to decrease the goodness in every hidden layer.

Following is the simplified version of Python implementation of the forward-forward algorithm and a schematic illustration.
\begin{figure}[h]
  \begin{subfigure}[b]{0.5\linewidth}
		\includegraphics[width=1.2\linewidth]{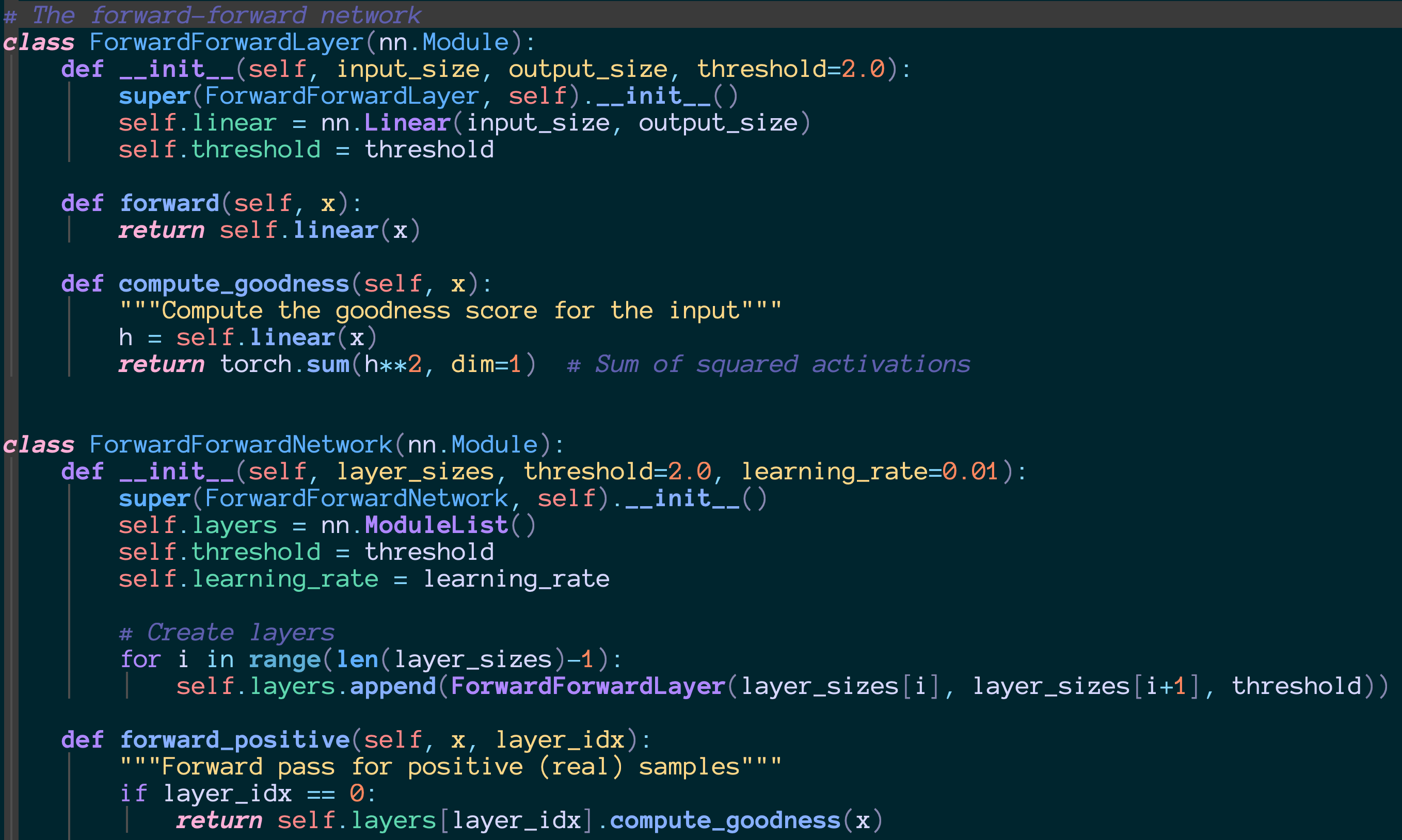}
	  \caption{A snippet of Python implementation of Forward-Forward layer}
	\end{subfigure}
	 \hfill
	\begin{subfigure}[b]{0.4\linewidth}
		\centering
		\includegraphics[width=1.1\linewidth]{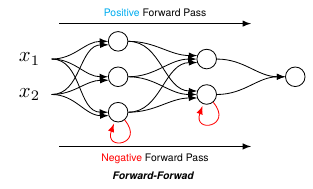}
	  \caption{Illustration}
	\end{subfigure}
	\caption{Forward-Forward Algorithm}
  \end{figure}

The transformed quantum version of the forward-forward algorithm is shown in the following figure. Noticed the implementation of the positive and negative pass is quantized with quantum circuits.

\begin{figure}[h]
  \centering
	\includegraphics[width=0.5\linewidth]{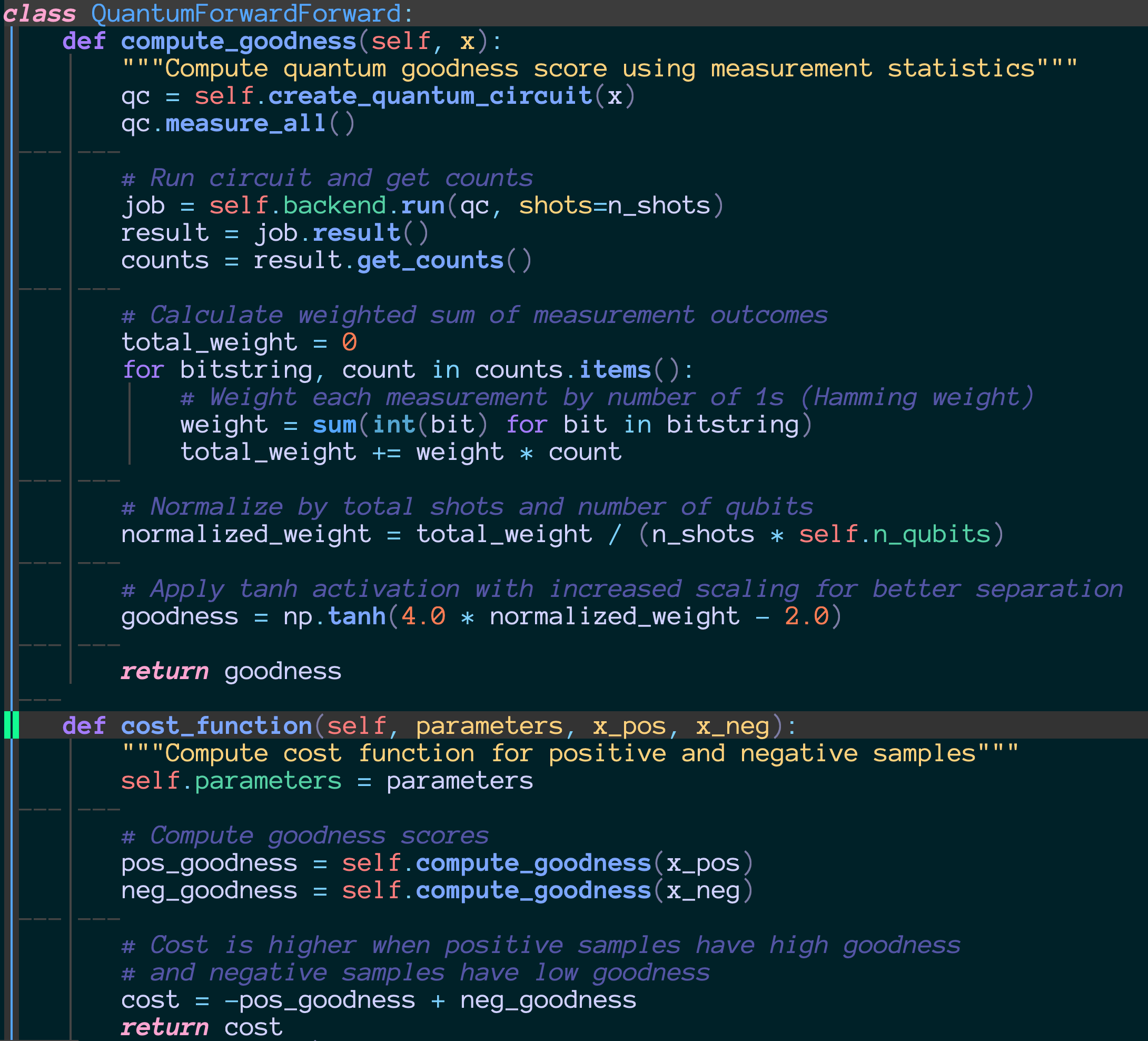}
	\caption{A snippet of simplified Quantum Forward-Forward layer}
  \label{fig:fig5}
\end{figure}

\subsection{Backpropagation Algorithm}
\begin{figure}[h]
  \begin{subfigure}[b]{0.5\linewidth}
		\includegraphics[width=1.2\linewidth]{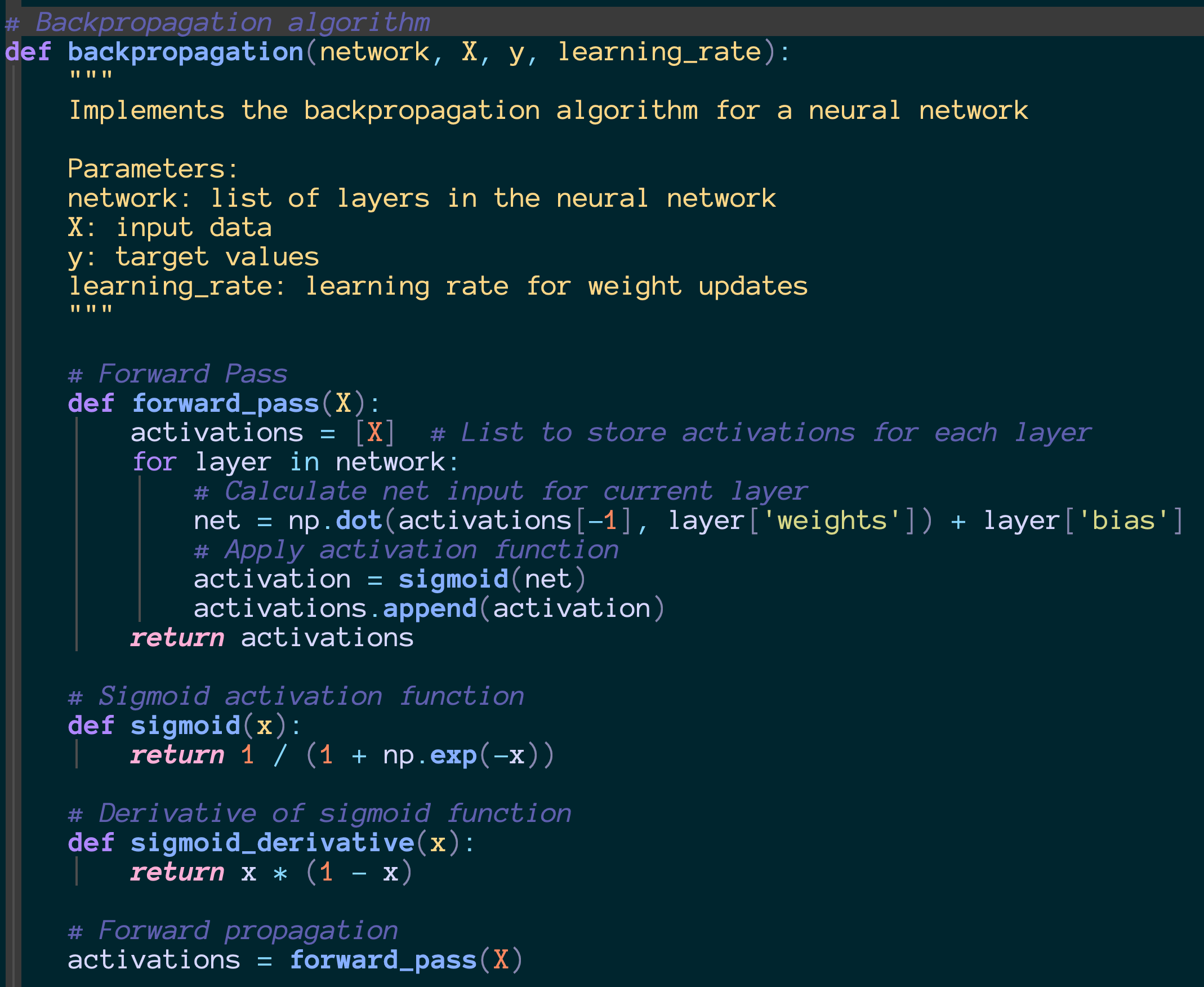}
	  \caption{A snippet of Python implementation of neural network}
	\end{subfigure}
	 \hfill
	\begin{subfigure}[b]{0.4\linewidth}
		\centering
		\includegraphics[width=1.1\linewidth]{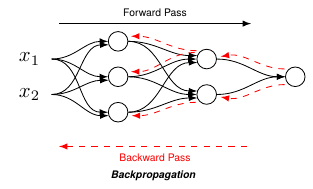}
	  \caption{Illustration}
	\end{subfigure}
	\caption{Backpropagation Algorithm}
  \label{fig:cbp}
\end{figure}

The backpropagation algorithm \parencite{rumelhartLearningRepresentationsBackpropagating1986} is probably the most fundamental optimization algorithm for deep neural networks. It is also more generalizable than the forward-forward algorithm, which is intended for networks where weight-sharing is not feasible. The backpropagation algorithm is a special case of the chain rule applied to neural networks. Through the efficient reuse of intermediate information, it facilitates network weight updating through gradient computation at a total cost approximately proportional to twice the runtime of the feedforward model (Figure \ref{fig:cbp}). Without incurring cost to an additional factor proportional to the number of parameters, it is effective in performing stochastic gradient descent with a large number of parameters and a lot of data, which brings the astonishing success of deep learning over the last decade. 

\begin{figure}[h]
  \centering
	\includegraphics[width=0.6\linewidth]{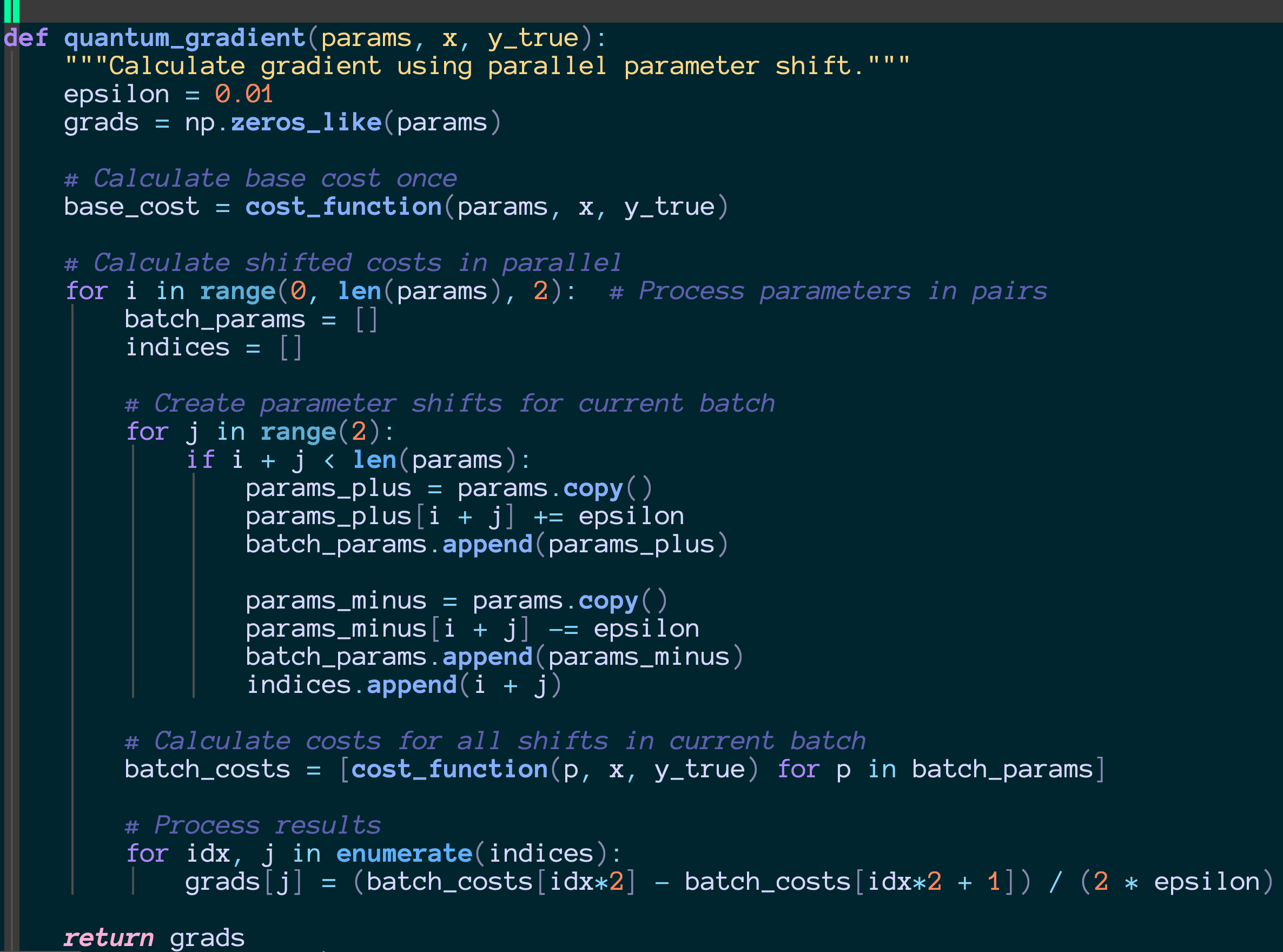}
	\caption{A snippet of the Quantum backpropagation with gradient}
  \label{fig:qbp}
\end{figure}

When specified with backpropagation algorithm, the agentic system would generate a quantum version of the backpropagation algorithm, which is shown in Figure \ref{fig:qbp}. The quantum version of the backpropagation algorithm is based on the quantum circuit model and uses quantum gates to perform the computation. And noticed that the gradient backpropagation was replaced with parameter-shift rule, which is a method for computing the gradient of a quantum circuit with respect to its parameters, and it is used to optimize the parameters of the quantum circuit.

\subsection{Performance Evaluation}
We compare the algorithms generated by the agentic system with the baseline human-crafted post-variational quantum neural network (from Pennylane Demos). The original demonstration employs a dataset from UCI ML hand-written digits datasets and pre-processes the dataset to make it a binary task, instead of original 10 classes. Here, the task was revert back to classification of the original 10 classes.   

We can spot that the quantum forward-forward algorithm got performance (average accuracy on the test set) similar to the baseline one (Table~\ref{tab:comparison}). Others are inferior to the baseline algorithm. Noted that we didn't training these model with massive training steps as the scope of this work is to demonstrate a proof of concept that our agentic framework can reach the ability to translate classical implementation of machine learn algorithms to its quantum counterpart. At the time when we run the agentic experiment, the base model was claude-3-5-sonnet-20240620.

\begin{table}[h]
\centering
\caption{\textbf{Comparison of Model Performance}}
\label{tab:comparison}
\begin{tabular}{|c|c|}
\hline
\textbf{Model} & \textbf{Average Accuracy (\%)} \\ \hline
Baseline (QNN) & 15.55 \\ \hline
QMLP & 9.40 \\ \hline
QFF & 15.17 \\ \hline
QBP & 12.37 \\ \hline
\end{tabular}
\end{table}

\section{Discussion}
Quantum-enhanced machine learning is an emerging field that explores the potential of quantum computing to improve machine learning algorithms. Quantum machine learning algorithms leverage the principles of quantum mechanics, such as superposition and entanglement, to perform computations that are infeasible for classical computers. The agentic framework we proposed can be used to search for quantum machine learning algorithms with aims to be more efficient and effective than classical counterparts. Since there are rich abundance of legacy classical machine learning concepts to explore, this primer work, as a proof of concept, focuses on the initial translation of the classical machine learning algorithms to its quantum counterpart. This operation shows the potential that allows us to adapt a plug and play searching strategy to automatically screen through classical machine learning concepts when developing quantum-enhanced machine learning algorithms. Thus, we consider the current agentic operation as a for-loop to iteratively search through machine learning concepts for the quantum building.

In abstraction, the operation can be abstracted as the following formulation.
\begin{align*}
  & \mathcal{A} =\sum_{ij} \alpha_{ij} \mathcal{L}_j \\
  & \mathcal{A}\left(\mathcal{C}_{ML}\right)=\argmin_\pi \mathcal{Q}_{ML}^\pi\\
\end{align*}

The LLMMA ($\mathcal{A}$) is composed by several large language models ($\mathcal{L}_j$; distinct agents, or lobes in the context of neural architecture) where $\alpha_{ij}$ specifies the interaction among the lobes. Given a classical machine learning algorithm ($\mathcal{C}_{ML}$), $\mathcal{A}$ can facilitate to find the corresponding quantum machine learning algorithm ($\mathcal{Q}_{ML}$) with optimized policy ($\pi$).

The work of AI Scientist \parencite{luAIScientistFully2024} has demonstrated an agentic operation to continually optimize on a template of code and ideas to find good performance machine learning model. Our work can be viewed as an extension that starts from a more primitive state of the initial condition, and perform transformation across knowledge domain. Then consider the transformative results as another session from which the agentic system operate on tasks of optimization and evolution. 

Currently, we are working on a variation that incorporates planning mechanism, either automatically generated serious of component to test through or synthesized with human guided, to run monte-carlo tree searching. The future direction of this work will be research on how to efficiently search through the space of classical machine learning concepts and translate them to quantum machine learning algorithms.

\printbibliography[title={References}]
\end{document}